\documentclass[runningheads,a4paper]{llncs}

\usepackage{amsmath, amssymb}
\usepackage{graphicx,enumitem, subfigure,url}

\begin{document}

\mainmatter  


\title{Near Real-time Hippocampus Segmentation Using Patch-based Canonical Neural Network}

\titlerunning{Near real-time hippocampus segmentation using PatchDNN}

%
%
 \author{Zhongliu Xie \hspace{2mm}  Duncan Gillies}

 \authorrunning{Zhongliu Xie and Duncan Gillies}

  \institute{Imperial College London \\ London, United Kingdom}
%
%

\maketitle

\begin{abstract}
Over the past decades, state-of-the-art medical image segmentation has heavily rested on signal processing paradigms, most notably registration-based label propagation and pair-wise patch comparison, which are generally slow despite a high segmentation accuracy. In recent years, deep learning has revolutionalized computer vision with many practices outperforming prior art, in particular the convolutional neural network (CNN) studies on image classification. Deep CNN has also started being applied to medical image segmentation lately, but generally involves long training and demanding memory requirements, achieving limited success. We propose a patch-based deep learning framework based on a revisit to the classic neural network model with substantial modernization, including the use of Rectified Linear Unit (ReLU) activation, dropout layers, 2.5D tri-planar patch multi-pathway settings. In a test application to hippocampus segmentation using 100 brain MR images from the ADNI database, our approach significantly outperformed prior art in terms of both segmentation accuracy and speed: scoring a median Dice score up to $90.98\%$ on a near real-time performance ($<1s$).

%
%
\end{abstract}

\section{Introduction}
Modern medical practices often involve the use of imaging technology (e.g. MRI) to examine pathology in vivo. Image segmentation is often a pre-requisite to further interventions. Due to the complexity and visual similarity of anatomical structures under medical imaging, state-of-the-art approaches often leverage a set of labeled training data, called atlases. Typically, the atlases are non-rigidly registered with a target image, and labels are propagated to perform segmentation by multi-atlas label propagation (MALP) \cite{MALP}. However, MALP generally requires a non-rigid registration between every atlas-target pair, which is time-consuming with a large atlas set. 
Other researchers have proposed a patch-based segmentation (PBS) framework \cite{Coupe,Rousseau}, in which the voxel-to-voxel correspondence for label propagation is relaxed to retrieve only matching patches from linearly aligned atlases for weighted label fusion. Patch similarity is typically measured using the sum of squared differences (SSD) in the form of pair-wise comparison. Since no non-rigid registration is required, it thus avoids the risk of registration failure and the need for pair-wise transformation, etc. Nevertheless, segmentation efficiency remains low, mostly due to the pair-wise patch comparison at run-time, although final speed may vary with different patch search methods. 
Furthermore, due to significant inter-subject variability, usually only a subset of atlases most similar to target image are used to improve performance in both frameworks, which raises the issue of atlas selection.

In recent years, deep learning has revolutionalized computer vision with many practices outperforming prior state-of-the-art, most notably the convolutional neural network (CNN) studies on image classification \cite{AlexNet}. In fact, early artificial neural network (ANN) research dates back to 1950s, yet has since achieved limited success, until modern GPU programming meets its massive computational requirement for network training. A neural network consists of a number of layers, each is composed by a set of independent neurons, and each neuron stores a feature function that takes an input vector and outputs a scalar value. With a deep structure (i.e. many layers), the aggregate feature function can be highly sophisticated and enables excellent classification capabilities. However, since an image may easily scale to millions of pixels, leading to millions of learning weights for a neuron and thousands of such neurons in a network, this architecture will quickly end up over-fitting. 
As a variant model,  
a CNN layer filters an input image with a sliding convolution kernel and outputs a multi-channel feature image (each channel corresponds to a neuron). The initial input is filtered layer-by-layer until an output layer that takes the aggregate features for classification. 
Deep CNN can also achieve sophisticated feature extraction while using far fewer learning weights, making it more robust and more popular than classic ANN for image classification. Although at a limited scale, deep CNN has started being applied to medical image segmentation lately \cite{CNN-pancreas,CNN-breast,2.5D}. Typically, segmentation is broken down to voxel-wise labeling on a patch-based setting, where a patch is treated as a mini-image for classification of its center voxel. 

However, voxel-wise label classification using deep CNNs generally involves thousands of neuron-wise convolution on each single run, leading to long training and demanding memory requirements (e.g. 5-6 days on two GPUs in \cite{AlexNet}). Moreover, existing CNN segmentation work is often based on a nesting structure that integrates with other models such as superpixels \cite{CNN-pancreas} or conditional random fields \cite{CNN-breast}, further complicating computation. By contrast in this paper we propose a patch-based deep neural network (patchDNN) based on a revisit to the classic ANN with substantial modern adaptations, in particular: a) the use of Rectified Linear Unit (ReLU) activation which speeds up training multiple times in other studies \cite{AlexNet}, b) deployment of dropout layers to address over-fitting \cite{dropout}, c) employment of 2.5D tri-planar patch multi-pathway setting \cite{2.5D} that secures a high accuracy at much lower computational cost and memory consumption than the conventional 3D setting. 
Although our approach is general, we conducted experimentation on hippocampus segmentation, and achieved a median Dice score up to 90.98$\%$ at a near real-time speed ($<$1s). To the best of our knowledge, this is the fastest algorithm with highest segmentation accuracy ever reported.


\section{Materials and Methods}

\subsection{Validation Database}
\begin{figure}[!t]
\centering
 \subfigure{\includegraphics[height=3cm]{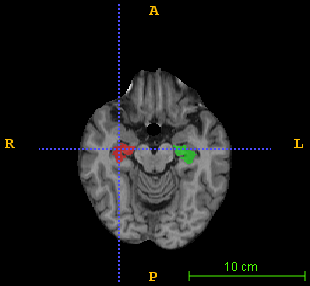}}
\hspace{0.2cm}
 \subfigure{\includegraphics[height=3cm]{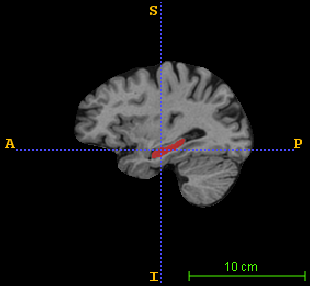}}
\hspace{0.2cm}
 \subfigure{\includegraphics[height=3cm]{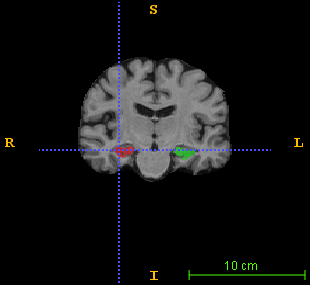}}
\caption{A sample brain MR image superimposed with its hippocampus reference segmentation in (left) axial view, (middle) sagittal view, and (right) coronal view. The green and pink colored regions respectively indicate the left and right hippocampus}
\label{fig:example}
\vspace{-0.5cm}
\end{figure}

A validation study has been conducted on the brain MRI data drawn from the Alzheimer's Disease Neuroimaging Initiative database (\url{http://adni.loni.usc.edu}). As a pilot study, 100 brain images were randomly picked: 34 healthy subjects, 33 with mild cognitive impairment (MCI) and 33 AD patients, with a demographic profile presenting no statistically significant difference on age and MMSE score from the whole database ($p-value > 0.1$). All images were bias-corrected and linearly aligned in the widely-used MNI152 template space. Each image also includes a reference manual segmentation of the hippocampus. A sample image superimposed with its label map is shown in Fig. \ref{fig:example}.

\subsection{Approach Overview}
Our approach carries out voxel-wise label classification using patch-based contextual information of each target voxel. In contrast to the time-consuming pair-wise patch comparison using low-level features (e.g. SSD) for label fusion in the prior art, we train a patch-based deep neural network (PatchDNN) that serves as both a sophisticated feature extraction mechanism and a classifier. The PatchDNN takes patch data as input and projects it to a high-level feature space through a set of deeply learned non-linear functions, followed by softmax classification. In this case, segmentation at run-time can become extremely fast and highly accurate. Furthermore, the 2.5D tri-planar patch setting \cite{2.5D} is borrowed, which takes three 2D patches respectively from the axial, coronal and sagittal planes as the input. It achieves comparable segmentation accuracy (sometimes better \cite{2.5D}) as the conventional 3D patch setting at a much smaller computational cost. In an abstract sense, the PatchDNN model can be formulated as:
\begin{equation}
F = h \left( P_{1}(v),P_{2}(v),P_{3}(v) \right),
\qquad 
L = softmax(F),
\qquad 
l = \underset{c}{argmax}  \left( L_{c} \right)
\label{eq1}
\end{equation}
where $P_{1}(v)$, $P_{2}(v)$, $P_{3}(v)$ are the tri-planar patches for target voxel $v$, $F$ is a feature vector generated by the feature extraction function $h(\cdot,\cdot,\cdot)$, $L$ is a vector of label values obtained by the softmax classifier, and $l$ is the output label, which is assigned to the label class $c$ carrying the highest label value $L_{c}$.

\subsection{Network Architecture}

\begin{figure}[!t]
\centering
\includegraphics[width=\textwidth]{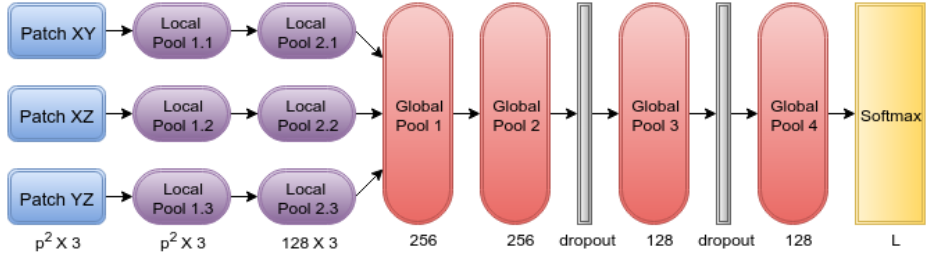}
\caption{The architecture of PatchDNN}
\label{fig:DNN}
\vspace{-0.5cm}
\end{figure}

The network contains 6 feature extraction layers and 2 dropout layers before a final softmax layer to output label values for classification, as shown in Fig. \ref{fig:DNN}. All layers preceding the softmax layer collectively model the $h(\cdot,\cdot,\cdot)$ in Eq. (\ref{eq1}). Each neuron in a feature extraction layer, pairing with a subsequent ReLU unit, independently generates a feature response by the following compound function:
\begin{equation}
f^{(k)} = \sum w_{i}^{(k)} \cdot x_{i} + b^{(k)}, \qquad \qquad g^{(k)}=max(f^{(k)},0)
\label{eq2}
\end{equation}
where $w_{i}^{(k)}$ and $b^{(k)}$ are learnable weights for the $i$th entry of (interim) input $x$ at $k$th neuron. The ReLU activation introduces non-saturating non-linearity, which was found able to speed up gradient descent training process multiple times than the traditional Sigmoid and Tanh activation functions \cite{AlexNet}. Furthermore, the first two feature extraction layers contain three pathways, respectively for each of the tri-planar patches. Starting the third layer, all three pathways merge into one, which becomes a standard fully connected layer. Such design significantly reduces the number of weights to train compared to a scheme with full connectivity from the beginning, and meanwhile secures a sophisticated level of feature representation. In addition, Layer 5 and 7 are dropout layers. A dropout layer randomly disconnects each neuron from the network at probability $\epsilon$, which is typically set to 0.5 during training and 0 (i.e. no dropout) at test time \cite{dropout}. The purpose of dropout is to simulate the network as a combination of several networks trained separately, which can empirically reduce over-fitting.

\subsection{Network Training}
Our framework performs a supervised learning scheme, where the patch tuple $<P_{1}(v)$, $P_{2}(v)$, $P_{3}(v)>$ collectively represents a single training data entry $d$ and the center voxel label in the reference segmentation indicates its supervisory output $l^{*}$. In terms of cost function, we resort to the widely-used cross entropy:
\begin{equation}
Min. \quad E = H(\Phi^{(j)}(X),Y)
\end{equation}
where $X=[d_{1},...,d_{n}]$, $Y=[l^{*}_{1},...,l^{*}_{n}]$ represent the training data and ground truth, while $\Phi^{(j)}(X)$ denotes PatchDNN output at training step $j$. Moreover, a number of leading-edge training techniques are employed in this work:

\begin{description}[style=unboxed,leftmargin=0cm,itemsep = 1mm]

\item[\textit{Pruning training pool with ROI mask:}] 
due to the structural similarity of human brains, hippocampi in different images locate spatially close (after linear alignment). We therefore apply a mask over the region of interest (ROI), created by taking the union of all labels in all atlases, followed with minor dilation. The mask was manually checked to ensure it fully covers each hippocampus and immediate neighborhood in each image. This can reduce training pool to a small fraction, dramatically lowering computational volume and memory consumption.

\item[\textit{Mini-batch stochastic gradient descent: }] 
moreover, the modern mini-batch gradient descent optimization scheme is employed. It takes a small batch (denote size $\delta$) of training data $X'$ and $Y'$, randomly sampled from the training pool, to perform an optimization step each time, instead of extending to the entire pool.

\item[\textit{Foreground/background separate sampling:}] background labels often significantly outnumber foreground labels, even within an ROI mask. In our case, the ratio was over 10:1. Such label imbalance could lead to a trained PatchDNN over-fitting the background label class. To address the issue, for each mini-batch we draw half the samples from foreground and half from background. This enables each foreground data entry to be trained multiple times of a background counterpart, and was proven able to effectively improve segmentation accuracy. 

\end{description}

\section{Results}

\subsection{Experimental setting and evaluation method}
\label{results}
Experimentation was carried out using cross-validation. The 100 images in the test database were divided into ten equally-sized folds via random distribution. A leave-one-out strategy was applied in each experiment, with nine folds to train a network and the remaining fold for testing. In total, there were ten instances of network training and 100 instances of segmentation for each experimental setting, which is considered adequate for our proof-of-concept purpose. Segmentation accuracy was measured using the Dice score (a.k.a. the kappa or similarity index), which is a prevailing metric standard computed by  $Dice(A,B) = \frac{2|A\cap B|}{|A|+|B|}$, i.e. the number of matching labels between segmentation $A$ and ground truth $B$, divided by the total number of labels in both label maps.

\subsection{Training parameters}

 \begin{figure}[!t]
 \centering
  \subfigure{\includegraphics[height=5cm, width=6.05cm]{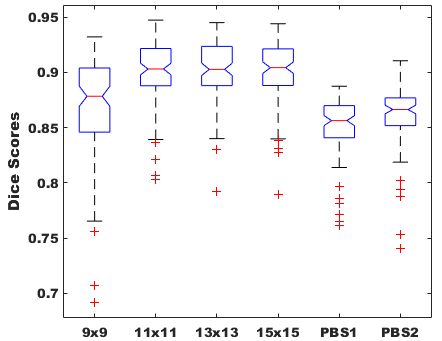}}
  \subfigure{\includegraphics[height=5cm, width=6.05cm]{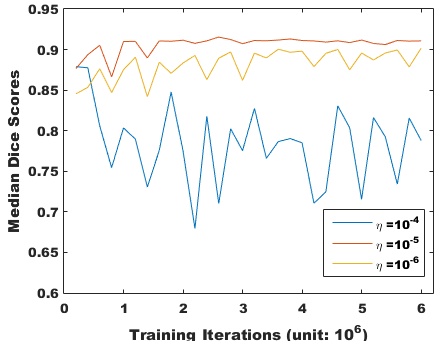}}
 \caption{Impact of training parameters: (left) patch size (right) learning rate $\eta$}
 \label{fig:parameter}
\vspace{-0.5cm}
 \end{figure}

\begin{figure}[!t]
\centering
\subfigure{
       \begin{minipage}[t]{0.3\textwidth}
		\centering
       \includegraphics[height = 2cm]{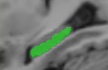}
		\text{\small Best case}
       \includegraphics[height = 2cm]{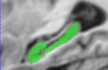}
		\text{\small Median case}
       \includegraphics[height = 2cm]{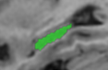}
		\text{\small Worst case}
       \end{minipage}
}
\subfigure{
       \begin{minipage}[t]{0.3\textwidth}
		\centering
       \includegraphics[height = 2cm]{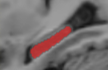}
		\text{\small Dice = 0.9464}
       \includegraphics[height = 2cm]{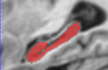}
		\text{\small Dice = 0.9098}
       \includegraphics[height = 2cm]{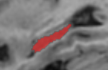}
		\text{\small Dice = 0.7930}
       \end{minipage}
}
\subfigure{
       \begin{minipage}[t]{0.3\textwidth}
		\centering
       \includegraphics[height = 2cm]{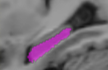}
		\text{\small Dice = 0.8315}
       \includegraphics[height = 2cm]{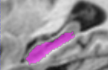}
		\text{\small Dice = 0.6226}
       \includegraphics[height = 2cm]{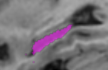}
		\text{\small Dice = 0.8153}
       \end{minipage}
}
\caption{Sample segmentation outcome: (left) reference segmentation, (middle) PatchDNN, and (right) prior state-of-the-art PBS method}
\vspace{-1em}
\label{fig:seg}
\end{figure}

In this study, there were around $70,000$ patch tuples extracted from each atlas, which amounted to 63 million data entries (90 atlases), collectively used to train a network with $O(10^{5})$ weights in each test. At training stage, the only major parameters to tune were: 1) patch size $p \times p$, 2) learning rate $\eta$, 3) mini-batch size $\delta$. For simplicity, $\delta$ was fixed at 200 entries for each training step. In terms of patch size, we tested four settings, respectively  $9 \times 9$, $11 \times 11$, $13 \times 13$ and $15 \times 15$. The performance metrics are shown in Fig. \ref{fig:parameter}(left). The highest median dice score achieved ($13 \times 13$ setting) was 90.98\%, which to the best of our knowledge, is by far the highest accuracy level ever reported on hippocampus segmentation. The $9 \times 9$ setting scored slightly lower than the others at 87.75\% level, yet still outperformed prior state-of-the-art (to be detailed in Section \ref{sec:PBS}). The learning rate is the tricky bit, and can dramatically affect training outcome. Figure \ref{fig:parameter} (right) illustrates the impact of three rates:  $\eta = 10^{-4}$, $\eta =  10^{-5}$ and $\eta =  10^{-6}$, on the accuracy of a PatchDNN using $13 \times 13$ patches: $\eta = 10^{-4}$ was too high and $\eta =  10^{-6}$ too low, whereas $\eta = 10^{-5}$ was considered the best rate in this case, which was also the rate used in our final setting. A sample segmentation outcome using a $13 \times 13$ PatchDNN trained 5 million times at $\eta = 10^{-5}$ is illustrated in Fig. \ref{fig:seg} (middle), in comparison to the reference segmentation (left).


\subsection{Training and testing time}
The experimental environment was deployed on a standard PC with an NVIDIA GTX Titan X graphics card. During training, each gradient descent step took  around 0.01$s$. Since it generally takes millions of steps to secure good performance, the total training time may take up to 10+ hours. In stark contrast, image segmentation at test time achieved a near real-time speed, taking $<1s$ for each target image. To the best of our knowledge, this is by far the fastest hippocampus segmentation system, with only the PatchMatch method \cite{patchmatch} having reported a comparable speed.


\subsection{Comparison with prior state-of-the-art: PBS framework}
\label{sec:PBS}
To further demonstrate the superiority of our framework, we also tested the PBS method \cite{Coupe} for direct comparison. The implementation was based on the built-in PBS framework in the open source IRTK repository (\url{https://github.com/BioMedIA/IRTK}). To ensure fairness, we rigorously applied the same pre-processing described in \cite{Coupe}, including nonlocal means denoising \cite{denoising}, N3 bias correction and tissue-standardizing normalization \cite{normalization}. For each target image, we then selected 10 atlases with lowest overall SSD within the masked ROI to perform segmentation. Two patch settings were tested: $5 \times 5 \times 5$ (PBS-1) and $7 \times 7 \times 7$ (PBS-2), over a search window $11 \times 11 \times 11$, which were arguably two best performing settings in \cite{Coupe}. The metrics are also shown in Fig. \ref{fig:parameter} (left) for easy comparison. The best median Dice score obtained was $86.65\%$, a level somewhat below the $88.4\%$ in the authors' own work, which might be caused by the use of different experimental datasets. However, by any means the performance was notably lower than our PatchDNN approach. A comparative segmentation outcome is illustrated in Fig. \ref{fig:seg}. Moreover, efficiency boost was even more evident, with PBS-1 and PBS-2 respectively taking an average $197s$ and $645s$ (excluding pre-processing) to segment a target image, compared to our near real-time level.


\section{Discussion and Conclusion}
Over the past decades, image segmentation in the biomedical domain in general has heavily rested on signal processing paradigms, with non-rigid registration (for MALP) and pair-wise patch comparison (for PBS) being the cornerstones of prior state-of-the-art. Although at a much limited scale, machine learning has been increasingly practiced in recent years. Machine learning approaches tend to be much faster than signal processing counterparts, however the level of sophistication for feature representation has been a limiting factor to secure a comparable accuracy using conventional frameworks such as random forests. Deep learning in contrast, by projecting contextual information over a series of nonlinear neuron-layers, is able to achieve highly sophisticated feature extraction and train a high performance classifier in terms of both accuracy and efficiency. 

Deep CNN is the model behind many groundbreaking image classification studies that overtook prior art in computer vision over recent years. Its major advantage for image classification over classic ANN is a comparatively small number of learning weights (although many systems still scale to millions, e.g. \cite{AlexNet}), whereas a fully connected ANN applied to a large image would end up with an overwhelming number of learning weights that easily lead to over-fitting. In the case of (medical) image segmentation however, we argue such concern in the patch-based (mini-image) setting is not as significant as at full image scale, and can be further relieved by the employment of modern techniques such as dropout layers and tri-planar patch multi-pathway setting. For that reason, we abandon the popular CNN model and propose PatchDNN, which can be considered a modernized classic ANN model. Without large-scale neuron-wise convolution on each single run, a PatchDNN trains much faster than a deep CNN while consuming far less memory. Moreover, advanced training techniques have also been utilized, in particular GPU programming, stochastic gradient descent and ReLU activation, which collectively reduce training time dramatically and make such an approach very practical. 
In a test application to hippocampus segmentation using 100 brain MR images from the ADNI database, our framework was able to significantly outperform the prior state-of-the-art PBS approach, in terms of both segmentation accuracy and speed: scoring a median Dice score up to $90.98\%$ with a near real-time performance ($<1s$) on a modern GPU.

%


\end{document}